\documentclass[11pt,a4paper]{article}
\usepackage[hyperref]{acl2018}
\usepackage{times}
\usepackage{latexsym}
\usepackage{tikz-dependency}
\usepackage{subfig}
\usepackage{url}

\aclfinalcopy 

\newlength{\eud}
\usepackage{enumitem}
\usepackage{caption}
\usepackage{graphicx}
\usepackage{lipsum}
\usepackage{multirow}
\usepackage{caption}
\usepackage{graphicx}
\usepackage{tabularx}
\usepackage{array}
\usepackage{booktabs}
\usepackage{nth}
\newcommand{\head}[1]{\textnormal{\textbf{#1}}}
\usepackage{amsmath}
\usepackage{textcomp}
\newcolumntype{P}[1]{>{\centering\arraybackslash}p{#1}}


\newcounter{eqnnosave}          
{\end{enumerate}%
\setcounter{eqnnosave}{\arabic{enumi}}%
}

\title{Syntactic Dependency Representations in Neural Relation Classification}

\author{Farhad Nooralahzadeh \and Lilja {\O}vrelid\\
         University of Oslo\\ Department of Informatics\\
         {\tt  \{farhadno,liljao\}@ifi.uio.no}}
\date{}
\hypersetup{draft}
\begin{document}
\maketitle
\begin{abstract}
 We investigate the use of different syntactic dependency representations in a neural relation classification task and compare the CoNLL, Stanford Basic and Universal Dependencies schemes. We further compare with a syntax-agnostic approach and perform an error analysis in order to gain a better understanding of the results.
\end{abstract}
\section{Introduction}
The neural advances in the field of NLP challenge long held assumptions regarding system architectures. The classical NLP systems, where components of increasing complexity are combined in a pipeline architecture are being challenged by end-to-end architectures that are trained on distributed word representations to directly produce different types of analyses traditionally assigned to downstream tasks. Syntactic parsing has been viewed as a crucial component for many tasks aimed at extracting various aspects of meaning from text, but recent work challenges many of these assumptions. For the task of semantic role labeling for instance, systems that make little or no use of syntactic information, have achieved state-of-the-art results \cite{Mar:Fro:Tit:17}. For tasks where syntactic information is still viewed as useful, a variety of new methods for the incorporation of syntactic information are employed, such as recursive models over parse trees \cite{Soc:Per:Wu:13,Ebrahimi2015ChainBR} , tree-structured attention mechanisms \cite{Kok:Pot:17},  multi-task learning \cite{Wu:Zha:Yan:17},  or the use of various types of syntactically aware input representations, such as embeddings  over syntactic dependency paths \cite{Xu:Mou:Li:15}.

Dependency representations have by now become widely used representations for syntactic analysis, often motivated by their usefulness in downstream application. There is currently a wide range of different types of dependency representations in use, which vary mainly in terms of choices concerning syntactic head status. 
Some previous studies have examined the
effects of dependency representations in various downstream applications \cite{Miy:Sae:Sag:08,Elm:Joh:Kle:13}. Most recently, the Shared Task on Extrinsic Parser Evaluation \cite{Oep:Ovr:Bjo:17} was aimed at
providing better estimates of the relative utility of different types
of dependency representations and syntactic parsers for downstream
applications. The downstream systems in this previous work have, however, been limited to traditional (non-neural) systems and there is still a need for a better understanding of the contribution of syntactic information in neural downstream systems.

In this paper, we examine the use of syntactic representations in a neural approach to the task of relation classification. We quantify the effect of syntax by comparing to a syntax-agnostic approach and further compare different syntactic dependency representations that are used to generate embeddings over dependency paths.
\begin{figure}[t]
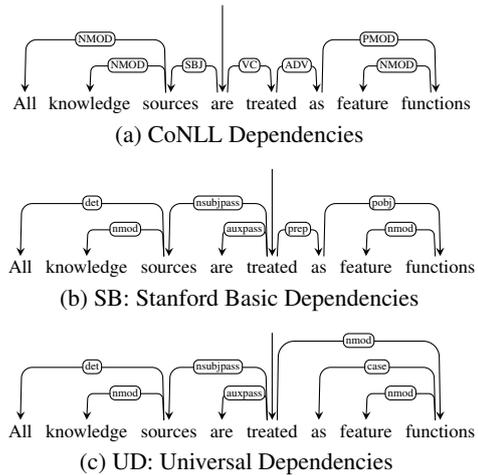

\centering\scriptsize
\captionsetup[subfloat]{farskip=1pt,captionskip=0.5pt}
\subfloat[CoNLL Dependencies\label{fg:conll08}]{
\begin{dependency}[edge below, edge slant=0.15ex,edge unit distance=\eud,edge horizontal padding=0.6ex]
\begin{deptext}[column sep=.05cm]
All \& knowledge \& sources \& are \& treated \& as \& feature \& functions\\
\end{deptext}
\depedge[edge above]{3}{1}{NMOD}
\depedge[edge above]{3}{2}{NMOD}
\depedge[edge above]{4}{3}{SBJ}
\depedge[edge above]{4}{5}{VC}
\depedge[edge above]{5}{6}{ADV}
\depedge[edge above]{6}{8}{PMOD}
\depedge[edge above]{8}{7}{NMOD}
\deproot[edge above,hide label]{4}{ROOT}
\end{dependency}}\\
\subfloat[SB:\ Stanford Basic Dependencies\label{fg:stanford}]{
  \begin{dependency}[edge below, edge slant=0.15ex,edge unit distance=\eud,edge horizontal padding=0.6ex]
  \begin{deptext}[column sep=.6ex]
All \& knowledge \& sources \& are \& treated \& as \& feature \& functions\\ 
  \end{deptext}
\depedge[edge above]{3}{1}{det}
\depedge[edge above]{3}{2}{nmod}
\depedge[edge above]{5}{3}{nsubjpass}
\depedge[edge above]{5}{4}{auxpass}
\depedge[edge above]{5}{6}{prep}
\depedge[edge above]{6}{8}{pobj}
\depedge[edge above]{8}{7}{nmod}
\deproot[edge above,hide label]{5}{root}
 \end{dependency}}\\
\subfloat[UD:\ Universal Dependencies \label{fg:ud}]{
  \begin{dependency}[edge below, edge slant=0.15ex,edge unit distance=\eud,edge horizontal padding=0.6ex]
  \begin{deptext}[column sep=.6ex]
All \& knowledge \& sources \& are \& treated \& as \& feature \& functions\\ 
  \end{deptext}
\depedge[edge above]{3}{1}{det}
\depedge[edge above]{3}{2}{nmod}
\depedge[edge above]{5}{3}{nsubjpass}
\depedge[edge above]{5}{4}{auxpass}
\depedge[edge above]{5}{8}{nmod}
\depedge[edge above]{8}{6}{case}
\depedge[edge above]{8}{7}{nmod}
\deproot[edge above,hide label]{5}{root}
  \end{dependency}}\\
\caption{Dependency representations for the example sentence.}
\label{fg:dep_graphs}
\end{figure}

\section{Dependency representations}
Figure \ref{fg:dep_graphs} illustrates the three different dependency representations we compare: the so-called CoNLL-style dependencies \cite{Joh:Nug:07} which were used for the 2007, 2008, and 2009 shared tasks of the
Conference on Natural Language Learning (CoNLL), the Stanford `basic' dependencies (SB) \cite{Mar:Mac:Man:06} and the Universal Dependencies (v1.3) (UD; \citealp{McD:Niv:Qui:13,Mar:Doz:Sil:14,Niv:Mar:Gin:16}). We see that the analyses differ both in terms of their choices of heads vs.\ dependents and the inventory of dependency types.
Where CoNLL analyses tend to view functional words as heads (e.g., the auxiliary verb \textit{are}), the Stanford
scheme capitalizes more on content words as heads (e.g., the main verb \textit{treated}). UD takes the tendency to select contentful heads one step further, analyzing the prepositional complement \textit{functions} as a head, with
the preposition {\it as} itself as a dependent case marker. This is in contrast to the CoNLL and Stanford scheme, where the preposition is head.

For syntactic parsing we employ the parser described in \newcite{Boh:Niv:12}, a transition-based parser which performs joint PoS-tagging and parsing. We train the parser on the standard training sections 02-21 of the Wall Street Journal (WSJ) portion of the Penn Treebank \cite{Mar:San:Mar:93}. The constituency-based treebank is converted to dependencies using two different conversion tools: (i) the pennconverter
software\footnote{\url{http://nlp.cs.lth.se/software/treebank-converter/}} \cite{Joh:Nug:07}, which produces the CoNLL dependencies\footnote{The pennconverter tool is run using the \texttt{rightBranching=false} flag.}, and (ii) the Stanford parser using either the option to produce basic dependencies \footnote{The Stanford parser is run using the \texttt{-basic} flag to produce the basic version of Stanford dependencies.} or its default option which is Universal Dependencies v1.3\footnote{Note, however, that the Stanford converter does not produce UD PoS-tags, but outputs native PTB tags.}.
The parser achieves a labeled accuracy score of 91.23 when trained on the CoNLL08 representation, 91.31 for the Stanford basic model and 90.81 for the UD representation, when evaluated against the standard evaluation set (section 23) of the WSJ. 
We acknowledge that these results are not  state-of-the-art parse results for English, however, the parser is straightforward to use and re-train with the different dependency representations. We also compare to another widely used parser, namely the pre-trained parsing model for English included in the Stanford CoreNLP toolkit \cite{manning-EtAl:2014:P14-5}, which outputs Universal Dependencies only. However, it was clearly outperformed by our version of the \newcite{Boh:Niv:12} parser in the initial development experiments.
\begin{figure*}[t]
\centering
\includegraphics[width=1\textwidth]{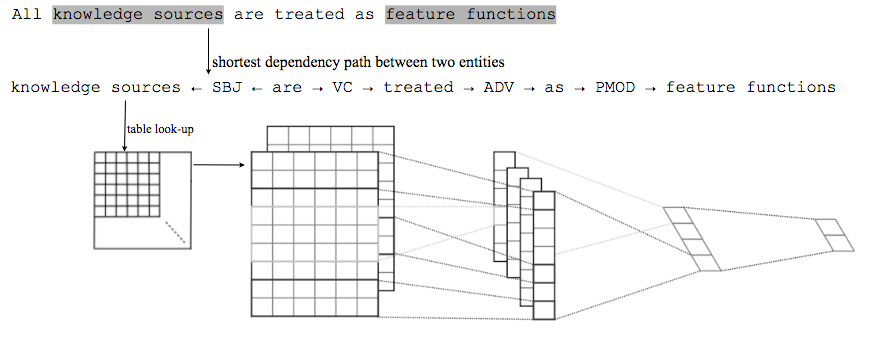}
\caption{Model architecture with two channels for an example shortest dependency path (CNN model from \newcite{DBLP:journals/corr/Kim14f}).}
\label{fig:cnn}
\end{figure*} 
\section{Relation extraction system}
We evaluate the relative utility of different types of dependency representations on the task of semantic relation extraction and classification in scientific papers, SemEval Task 7 \cite{SemEval2018Task7}. We make use of the system of \citet{Noo:Ovr:Lon:18}: a CNN classifier with dependency paths as input, which ranked 3rd (out of 28) participants in the overall evaluation of the shared task. Here, the shortest dependency path (\emph{sdp}) connecting two target entities for  each relation instance is provided by the parser and is embedded in the first layer of a CNN. We extend on their system by (i) implementing a syntax-agnostic approach, (ii) implementing hyper-parameter tuning for each dependency representation, and (iii) adding Universal Dependencies as input representation. We thus train classifiers with \emph{sdp}s extracted from the different dependency representations discussed above and measure the effect of this information by the performance of the classifier.

\subsection{Dataset and Evaluation Metrics} We use the SemEval-2018, Task 7 dataset \cite{SemEval2018Task7} from its \emph{Subtask 1.1}. The training data contains abstracts of 350 papers from the ACL Anthology Corpus, annotated for concepts and semantic relations. Given an abstract of a scientific paper with pre-annotated domain concepts, the task is to perform relation classification.
The classification sub-task 1.1 contains $1228$ entity pairs that are annotated based on five asymmetric relations ({\small USAGE, RESULT, MODEL-FEATURE, PART\_WHOLE, TOPIC}) and one symmetric relation ({\small COMPARE}). The relation instance along with its directionality are provided in both the training and the test data sets. 
The official evaluation metric is the macro-averaged F1-scores for the six semantic relations, therefore we will compare the impact of different dependency representations on the macro-averaged F1-scores.

The training set for Subtask 1.1 is quite small, which is a challenge for end-to-end neural methods. To overcome this, we combined the provided datasets  for Subtask 1.1 and Subtask 1.2 (relation classification on noisy data), which provides additional 350 abstracts and 1248 labeled entity pairs to train our model. This yields a positive impact ($+16.00\%$ F1) on the classification task in our initial experiments.
\subsection{Pre-processing}
Sentence and token boundaries are detected using the Stanford CoreNLP tool \cite{manning-EtAl:2014:P14-5}. Since most of the entities are multi-word units, we replace the entities with their codes in order to obtain a precise dependency path. Our example sentence {\it All knowledge sources are treated as feature functions}, an example of the {\small USAGE} relation between the two entities \textit{knowledge sources} and \textit{feature functions}, is thus transformed to {\tt All P05\_1057\_3 are treated as P05\_1057\_4}.

Given an encoded sentence, we find the \emph{sdp} connecting two target entities for each relation instance using a syntactic parser.
Based on the dependency graph output by the parser, we extract the shortest dependency path connecting two entities. The path records the direction of arc traversal using left and right arrows (i.e. $\leftarrow$ and $\rightarrow$) as well as the dependency relation of the traversed arcs and the predicates involved, following \newcite{DBLP:journals/corr/XuFHZ15}. The entity codes in the final \emph{sdp} are replaced with the corresponding word tokens at the end of the pre-processing step. For the sentence above, we thus extract the path: \texttt{knowledge sources $\leftarrow$ SBJ $\leftarrow$ are $\rightarrow$ VC $\rightarrow$ treated $\rightarrow$ ADV $\rightarrow$ as $\rightarrow$ PMOD $\rightarrow$ feature functions}

\subsection{CNN model} The system is based on a CNN architecture similar to the one used for sentence classification in \newcite{DBLP:journals/corr/Kim14f}. 
Figure ~\ref{fig:cnn} provides an overview of the proposed model. 
It consists of 4 main layers as follows: 1) {\bf Look-up Table and Embedding layer:} In the first step, the model takes a shortest dependency path (i.e., the words, dependency edge directions and dependency labels) between entity pairs as input and maps it into a feature vector using a look-up table operation. Each element of the dependency path (i.e. word, dependency label and arrow) is transformed into a embedding layer by looking up the embedding matrix $ M \in \mathcal{R}^{d\times V} $, where $d$ is the dimension of CNN embedding layer and $V$ is the size of vocabulary. Each column in the embedding matrix can be initialized randomly or with pre-trained embeddings. The dependency labels and edge directions are always initialized randomly.  2) {\bf Convolutional Layer}: The next layer performs convolutions with ReLU activation over the embeddings using multiple filter sizes and extracts feature maps. 3) {\bf Max pooling Layer}: By applying the \emph{max} operator, the most effective local features are generated from each feature map. 4) {\bf Fully connected Layer}: Finally, the higher level syntactic features are fed to a fully connected \emph{softmax} layer which outputs the probability distribution over each relation.
\begin{table*}[t]
 \setcounter{table}{1}
  \centering
  \scalebox{.8}{
  \begin{tabular}{*{9}{lP{1cm}P{3cm}P{1.5cm}cP{1.5cm}P{1.5cm}P{2cm}P{2cm}}}
  \toprule
    &\multicolumn{6}{c}{\head{Hyper parameters}}&\multicolumn{2}{c}{\head{F1}.(avg. in 5-fold)}\\
    \cmidrule(r){2-7}
    \cmidrule(r){8-9}
    \head{Representation} & Filter size & Num. Feature maps& Activation func.& L2 Reg.& Learning rate& Dropout Prob.& with default values& with optimal values \\
    \midrule
    CoNLL08  & 4-5 & 1000 & Softplus & 1.15e+01 & 1.13e-03 & 1 & 73.34  & 74.49\\
     SB  & 4-5 & 806 & Sigmoid & 8.13e-02 & 1.79e-03 & 0.87 & 72.83  & {\bf 75.05}\\
      UD v1.3  &5 & 716 & Softplus & 1.66e+00 &  9.63E-04 & 1 & 68.93 & 69.57 \\
    \bottomrule
  \end{tabular}
  }
   \caption{Hyper parameter optimization results for each model with different representation. The $max$ pooling strategy consistently performs better in all model variations.}
   \label{hyperparam}
\end{table*}
\section{Experiments}
We run all the experiments with a multi-channel setting\footnote{Initial experiments show that the multi-channel model works better than the single channel model} in which the first channel is initialized with pre-trained embeddings \footnote{We train 300-d domain-specific embeddings on the ACL Anthology corpus using the available word2vec implementation \emph{gensim} for training.} in static mode (i.e. it is not updated during training) and the second one is initialized randomly and is fine-tuned during training (non-static mode). The macro F1-score is measured by 5-fold cross validation and to deal with the effects of class imbalance,  we weight the cost by the ratio of class instances, thus each observation receives a weight, depending on the class it belongs to. 

\begin{table}
 \setcounter{table}{0}
  \centering
    \scalebox{.87}{
  \begin{tabular}{*{4}{lrrr}}
    \toprule
    & \multicolumn{2}{c}{\head{best F1} (in 5-fold)} & \\
    \cmidrule(ll){2-3}
    \head{Relation} & without sdp  & with sdp  & \head{Diff.} \\
    \midrule
    {\small USAGE} &60.34	&80.24	&+ 19.90  \\
     {\small MODEL-FEATURE} & 48.89 &	70.00	& + 21.11  \\ 
         {\small PART\_WHOLE}& 29.51&	70.27&	+40.76 \\
         {\small TOPIC} &45.80 &	91.26	& +45.46 \\
          {\small RESULT}& 54.35	& 81.58	& +27.23  \\
         {\small COMPARE} & 20.00  &61.82&	+ 41.82 \\
    \bottomrule
     macro-averaged & 50.10 &  76.10  & +26.00 \\
  \end{tabular}}
   \caption{Effect of using the shortest dependency path on each relation type. }
   \label{tbl.sdp}
\end{table}
\begin{table*}[t]
\setcounter{table}{3}
  \centering
    \scalebox{.7}{
  \begin{tabular}{lcr}
    \toprule
    \head{Sentence} & \multicolumn{1}{l}{\small
    This indicates that there is no need to add {\color{white}\colorbox{gray}{punctuation}} in transcribing {\color{white}\colorbox{gray}{spoken corpora}} simply in order to help parsers.} & class: {\small PART\_WHOLE} \\
    \midrule
      {\small CoNLL08} & {\small punctuation $\leftarrow$ obj $\leftarrow$ add $\rightarrow$ adv $\rightarrow$ in $\rightarrow$ pmod $\rightarrow$ transcribing $\rightarrow$ obj $\rightarrow$ spoken corpora} &\\
        {\small SB} &{\small punctuation $\leftarrow$ dobj $\leftarrow$ add $\rightarrow$ prep $\rightarrow$ in $\rightarrow$ pcomp $\rightarrow$ transcribing $\rightarrow$ dobj $\rightarrow$ spoken corpora} & \\  
         {\small UD v1.3}& {\small punctuation  $\leftarrow$ dobj  $\leftarrow$ add $\rightarrow$ advcl $\rightarrow$ transcribing $\rightarrow$ dobj $\rightarrow$ spoken corpora} & \\
    \midrule
    \midrule
    {\head{Sentence}} & \multicolumn{1}{l} {\small In the process we also provide a {\color{white}\colorbox{gray}{formal definition}} of {\color{white}\colorbox{gray}{parsing}} motivated by an informal notion due to Lang .} & class: {\small MODEL-FEATURE}\\
    \midrule
    {\small CoNLL08} & {\small formal definition $\rightarrow$ nmod $\rightarrow$ of $\rightarrow$ pmod $\rightarrow$ parsing} &\\
    {\small SB} &{\small formal definition $\rightarrow$ prep $\rightarrow$ of $\rightarrow$ pobj $\rightarrow$ parsing}&\\
    {\small UD v1.3}& {\small formal definition $\rightarrow$ nmod $\rightarrow$ parsing}&\\
    \midrule
    \midrule
    \head{Sentence} & \multicolumn{1}{l}{\small This paper describes a practical {\color{white}\colorbox{gray}{"black-box" methodology}} for automatic evaluation of {\color{white}\colorbox{gray}{question-answering NL systems}} in spoken dialogue.} & class: {\small USAGE}\\
    \midrule
    {\small CoNLL} & {\small " "black-box" methodology $\rightarrow$ nmod $\rightarrow$ for $\rightarrow$ pmod $\rightarrow$ evaluation $\rightarrow$ nmod $\rightarrow$ of $\rightarrow$ pmod $\rightarrow$ question-answering NL systems} &\\
    {\small SB} &{\small "black-box" methodology $\rightarrow$ prep $\rightarrow$ for $\rightarrow$ pobj $\rightarrow$ evaluation $\rightarrow$ prep $\rightarrow$ of $\rightarrow$ pobj $\rightarrow$ question-answering NL systems} & \\
    {\small UD v1.3}& {\small "black-box" methodology $\rightarrow$ nmod $\rightarrow$ evaluation $\rightarrow$ nmod $\rightarrow$ question-answering NL systems} & \\
    \bottomrule
  \end{tabular}}
   \caption{The  examples for which the CoNLL/SB-based models correctly predict the relation type in 5-fold trials, whereas the UD based model has an incorrect prediction.}
   \label{tbl.examples}
\end{table*}
\subsection{Effect of syntactic information}
To evaluate the effects of syntactic information in general for the relation classification task, we compare the performance of the model with and without the dependency paths. In the syntax-agnostic setup, a sentence that contains the participant entities is used as input for the CNN. We keep the value of hyper-parameters equal to the ones that are reported in the original work \cite{DBLP:journals/corr/Kim14f}. To provide the \emph{sdp} for the syntax-aware version we compare to, we use our parser with Stanford dependencies.
We find that the effect of syntactic structure varies between the different relation types. However, the \emph{sdp} information has a clear positive impact on all the relation types (Table \ref{tbl.sdp}). It can be attributed to the fact that the context-based representations suffer from irrelevant sub-sequences or clauses when target entities occur far from each other or there are other target entities in the same sentence. The \emph{sdp} between two entities in the dependency graph captures a condensed representation of the information required  to assert a relationship between two entities \cite{Bunescu:2005:SPD:1220575.1220666}.

\subsection{Comparison of different dependency representations} To investigate the model performance with various parser representations, we create a \emph{sdp} for each training example using the different parse models and exploit them as input to the relation classification model. With the use of default parameters there is a chance that these favour one of the representations. In order to perform a fair comparison, we make use of Bayesian optimization \cite{DBLP:journals/corr/abs-1012-2599} in order to locate optimal hyper parameters for each of the dependency representations. We construct a Bayesian optimization procedure using a Gaussian process with 100 iterations and Expected Improvement (EI) for its acquisition functions. We set the objective function to maximize the macro F1 score over 5-fold cross validation on the training set. Here we investigate the impact of various system design choices with the following parameters:  \footnote{Default values are \{3-4-5,  128,  ReLU, max,  3,  1e-3, 0.5\}}: I) Filter region size: $\in$ \{3, 4, 5, 6, 7, 8, 9, 3-4, 4-5, 5-6, 6-7, 7-8, 8-9, 3-4-5, 4-5-6, 5-6-7, 6-7-8, 7-8-9\} II) Number of feature maps for each filter region size: $\in \{10: 1000\}$ III) Activation function: $\in \{Sigmoid, ReLU, Tanh, Softplus, Iden \}$. IV) Pooling strategy: $ \in \{max, avg\}$. V) L2 regularization: $ \in \{1e-4: 1e+2\}$. VI) Learning rate: $\in \{1e-6: 1e-2\}$. VII) Dropout probability \footnote{The probability that each element is kept, in which $1$ implies that none of the nodes are dropped out}: $\in \{0.1: 1\}$. Table \ref{hyperparam} presents the optimal values for each configuration using different dependency representations. We see that the optimized parameter settings vary for the different representations, showing the importance of tuning for these types of comparisons. The results furthermore show that the \emph{sdp}s based on the Stanford Basic (SB) representation provide the best performance, followed by the CoNLL08 representation. We observe that the results for the UD representation are quite a bit lower than the two others.
\section{Error analysis}
Table \ref{tbl.repsf1} presents the effect of each parser representation in the classification task, broken down by relation type. We find that the UD-based model falls behind the others on the most relation types (i.e, {\small COMPARE, MODEL-FEATURE, PART\_WHOLE, TOPICS}).  
To explore these differences in more detail, we manually inspect the instances for which the CoNLL/SB-based models correctly predict the relation type in 5-fold trials, whereas the UD-based model has an incorrect prediction.
\begin{table}[!htpb]
\setcounter{table}{2}
  \centering
    \scalebox{.9}{
  \begin{tabular}{*{5}{lcrrr}}
    \toprule
    & &\multicolumn{3}{c}{\head{best F1} (in 5-fold) } & \\
    \cmidrule(ll){3-5}
    \head{Relation} & \head{Frq.}& CoNLL  & SB  & UD \\
    \midrule
    {\small USAGE}&947 & 76.84 & 82.39  & 77.56    \\
   
    {\small MODEL-FEATURE}&498 &  68.27  &  68.54   & 66.36  \\  
    {\small PART\_WHOLE}&425&  75.32 & 71.28  &    67.11\\
     {\small TOPIC} &258& 89.32 & 90.57  &    87.62\\
    {\small RESULT}&193 &82.35 & 81.69 & 82.86   \\
     {\small COMPARE} & 136& 66.67  & 66.67 &   54.24\\
 
    \bottomrule
    macro-averaged && 76.94 & 77.57  & 72.83
  \end{tabular}}
   \caption{Effect of using the different parser representation on each relation type.}
   \label{tbl.repsf1}
\end{table}
Table \ref{tbl.examples} shows some of these examples, marking the entities and the gold class of each instance and also showing the \emph{sdp} from each representation. We observe that the UD paths are generally shorter. A striking similarity between most of the instances is the fact that one of the entities resides in a prepositional phrase. Whereas the SB and CoNLL paths explicitly represent the preposition in the path, the UD representation does not. Clearly, the difference between for instance the {\small USAGE} and {\small PART\_WHOLE} relation may be indicated by the presence of a specific preposition ({\it X for Y} vs. {\it X of Y}). This is also interesting since this particular syntactic choice has been shown in previous work to have a negative effect on intrinsic parsing results for English \cite{Sch:Abe:Rap:12}.

\section{Conclusion}
This paper has examined the use of dependency representations for neural relation classification and has compared three widely used representations. We find that representation matters and that certain choices have clear consequences in downstream processing. Future work will extend the study to neural dependency parsers and other relation classification data sets.

\bibliography{acl2018}
\bibliographystyle{acl_natbib}

\end{document}